\documentclass[letterpaper, 10 pt, conference]{ieeeconf}  

\usepackage{amsmath}
\usepackage{amssymb}
\usepackage{graphicx}
\usepackage{booktabs}
 \usepackage{multirow}
 \usepackage{hyperref}
 \usepackage{xcolor}
 \usepackage[english]{babel}

\IEEEoverridecommandlockouts                              

\title{\LARGE \bf
SearchDiffusion: Information-Guided Diffusion Planning for 

Object Search
}

\author{Sausar Karaf, Dmitrii Plotnikov, Alexey Kotcov, and Dzmitry Tsetserukou%
\thanks{The authors are with the Intelligent Space Robotics Laboratory, Skolkovo Institute of Science and Technology, Bolshoy Boulevard 30, bld. 1, 121205, Moscow, Russia
\tt \{Sausar.Karaf, Dmitrii.Plotmikov, Alexey.Kotcov, D.Tsetserukou\}@skoltech.ru}}

\begin{document}

\maketitle


\thispagestyle{empty}
\pagestyle{empty}

\begin{abstract}
Object search in unknown environments requires robots to gather useful observations while avoiding obstacles and accounting for limited sensor visibility. We present SearchDiffusion, a diffusion-based planner that generates position and heading trajectories using a learned dense utility map. The map represents geometric accessibility, useful visibility, and semantic progress. It conditions trajectory generation and provides an objective for sampling guidance. Sampled map hypotheses provide additional geometric context, while geometric and heading guidance, validity checks, and analytical selection support execution. Across 90 simulated tasks on 24 map layouts, SearchDiffusion achieved 95.6\% target detection at greater computational cost. Ablations show complementary benefits from utility and guidance, with limited additional detection benefit from runtime map hypotheses. In ten physical quadruped trials using OneMap relevance maps, eight were successful under operator assessment, without fine-tuning on real-robot data. These results support combining learned spatial utility with guided trajectory generation for object search under partial observability.
\end{abstract}

\section{INTRODUCTION}

Autonomous object search requires robots to gather useful observations while avoiding obstacles. Observation quality depends on the camera's field of view, range, and viewing direction. Informative path planning therefore considers expected observation value alongside motion and safety constraints.

Diffusion planners learn trajectory distributions and generate candidates through iterative denoising~\cite{ho2020denoising,janner2022diffuser}. Object search requires adaptation to new geometric and semantic evidence. Semantic relevance alone does not indicate accessibility or routes that provide useful views, motivating a spatial utility map that relates available evidence to search progress.

We propose \textbf{SearchDiffusion}, a trajectory planner that learns dense utility from geometric accessibility, useful visibility, and semantic relevance. The predicted map conditions trajectory denoising and provides a differentiable objective for sampling guidance.

\begin{figure}[t!]
\centering
\includegraphics[width=0.7\columnwidth]{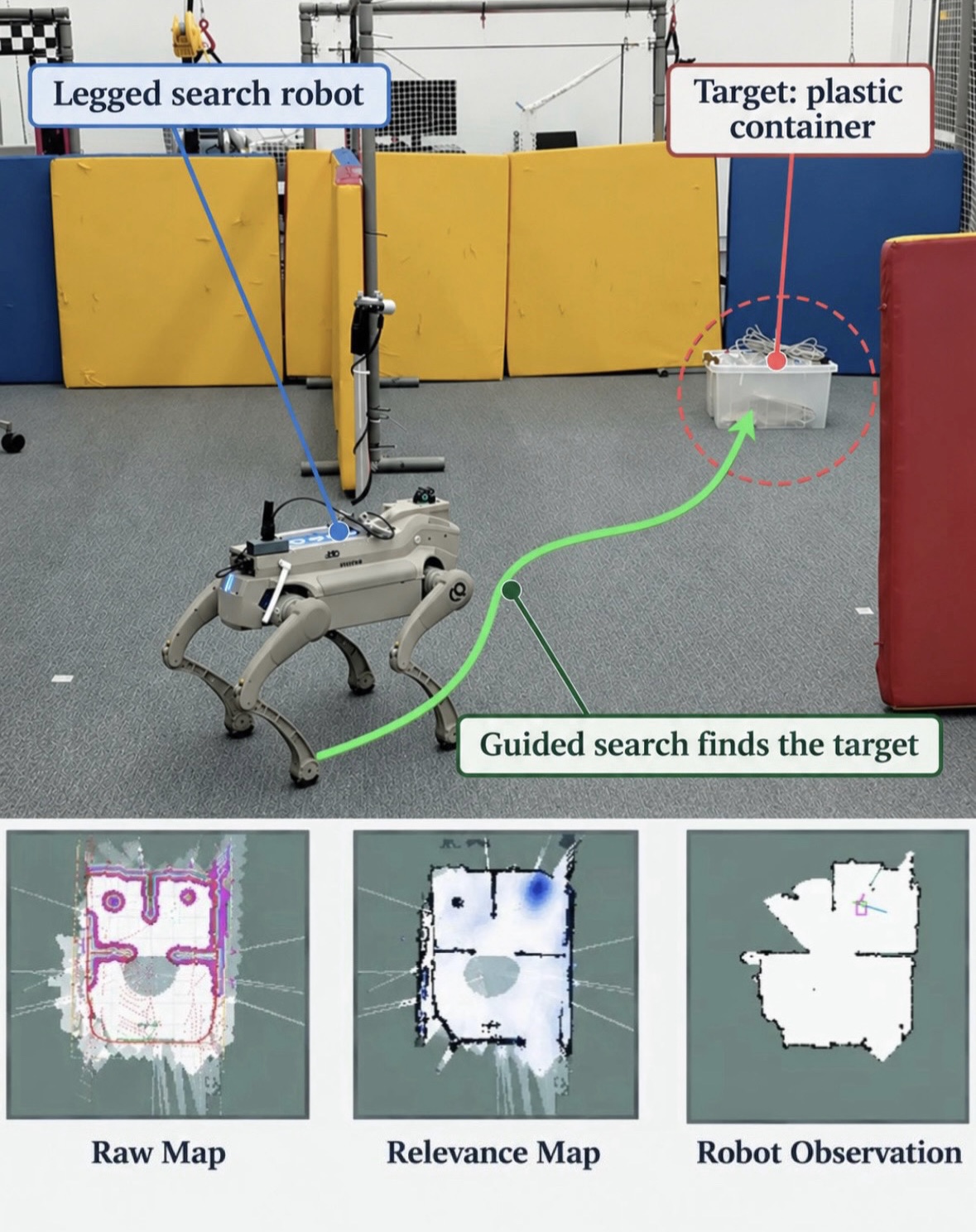}
\caption{SearchDiffusion for object search in a partially observed environment.}
\label{fig:searchdiffusion}
\end{figure}

\begin{figure*}[!htbp]
\centering
\includegraphics[width=0.85\textwidth]{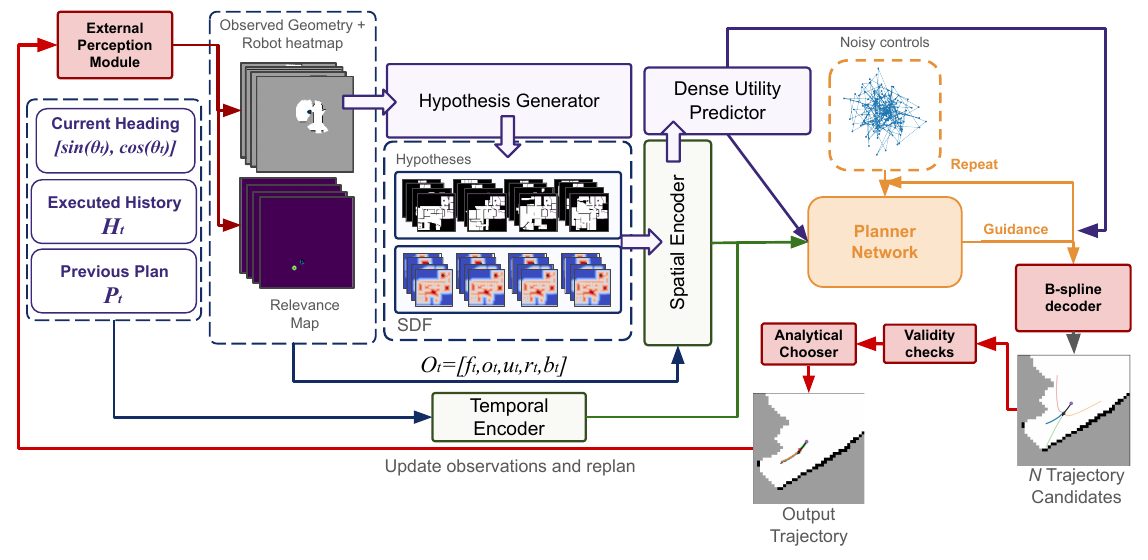}
\caption{SearchDiffusion overview. Predicted utility conditions trajectory denoising and guides sampling. Geometric and heading guidance refine candidates, followed by validity checks and analytical selection. The robot executes a trajectory prefix before replanning.}
\label{fig:SystemOverview}
\end{figure*}

Counterfactual map hypotheses provide context about unobserved geometry while preserving observed free and occupied cells. Utility, geometric, and heading guidance refine candidates before validity checks and analytical ranking. The robot executes an initial portion of the selected trajectory, updates observations, and replans.

We compare detection, trajectory efficiency, and planning time under matched simulated conditions. Ablations disable inference components with network weights fixed. Physical robot experiments use OneMap~\cite{Busch2025onemap} relevance maps to demonstrate compatibility with perception-derived inputs.

The main \textbf{contributions} are:
\begin{itemize}
\item \textbf{Utility-conditioned trajectory generation.}
A learned dense utility map encodes geometric accessibility, useful visibility, and semantic progress to condition trajectory diffusion.
\item \textbf{Guidance, validation, and selection.}
Utility, geometric, and heading objectives refine candidates during sampling. Validity checks and analytical scoring select a trajectory for partial execution.
\item \textbf{Evaluation of component interactions.}
Across 90 simulated tasks in 24 map clusters, SearchDiffusion improves detection over the tested baselines. Ablations support combining utility and guidance, with limited additional detection benefit from runtime hypotheses.
\end{itemize}

\section{RELATED WORK}

Classical exploration methods plan from the robot's current map. Yamauchi~\cite{yamauchi1997} introduced frontier-based exploration toward boundaries between known free space and unexplored regions. FASTER~\cite{tordesillas2022faster} plans through known and unknown space while maintaining a stopping backup in known free space. EGO-Planner~\cite{Zhou2021egoplanner} derives obstacle-avoidance gradients without a Euclidean signed distance field. Hollinger and Sukhatme~\cite{hollinger2014sampling} combine sampling-based search and pruning to maximize information quality under motion and budget constraints. Schmid et al.~\cite{Schmid2020} expand and rewire a trajectory tree to balance information gain and motion cost.

Language and vision models connect semantic information to robot behavior. FlockGPT~\cite{Lykov2024FlockGPTGU} translates instructions into UAV formations executed by a flocking controller. UAV-VLA~\cite{Sautenkov2025UAVVLAVS} combines language models and satellite imagery to generate flight paths and action plans. GoalSwarm~\cite{james2026goal} combines open-vocabulary perception, shared semantic mapping, and decentralized frontier allocation for multi-UAV object navigation.

Diffusion planners model multimodal trajectory distributions. Janner et al.~\cite{janner2022diffuser} introduced trajectory denoising with goal conditioning and reward guidance. Applications include autonomous driving~\cite{zheng2025diffusionbased} and quadruped path planning~\cite{Stamatopoulou2024DiPPeSTDP}. EmbodiedDiffusion~\cite{zhura2026embodieddiffusion} conditions generation on predicted dense traversability and uses a traversability-based training objective across robot platforms.

For exploration, DARE~\cite{cao2024dare} conditions diffusion on an attention-encoded graph of observed free space and frontier utilities. It learns from complete-map demonstrations but uses partial observations during deployment. GUIDE~\cite{che2025guide} adds LaMa-predicted free-space nodes, filtered by proximity to the robot and observed frontiers, to condition trajectory generation.

Other methods predict unseen environments for planning. MapEx~\cite{ho2024mapex} combines ensemble LaMa predictions, probabilistic visibility, and travel distance to select frontiers, then plans with A*. For goal-directed navigation, BeliefDiffusion~\cite{quilter2026beliefdiffusion} samples occupancy maps with diffusion, generates A* proposals on individual samples, and evaluates them using an aggregated map. Schr{"o}dinger's Navigator~\cite{he2025schrodinger} generates trajectory-conditioned 3D futures and combines their semantic, exploration, and risk cues into a value map, used alongside observed information for waypoint selection.

SearchDiffusion uses learned dense search utility to condition trajectory denoising and guide sampling. Observation-consistent map hypotheses provide geometric context, while validity checks and analytical selection support execution.

\section{PROBLEM FORMULATION}

We consider object search in a partially observed $M\times M$ grid
under limited sensing range and field of view. At planning step $t$,
the robot pose is $q_t=(x_t,y_t,\theta_t)$, where $\theta_t$ denotes
body yaw and determines the viewing direction through a fixed sensor
mounting. Position and body heading are planned jointly. Heading may differ from
the direction of translation; guidance encourages alignment through
a soft penalty rather than a hard constraint.

The robot accumulates observations into disjoint binary known-free and known-occupied masks, $f_t$ and $o_t$, starting with the observations available at initialization. Their complement $u_t=1-f_t-o_t$ marks unknown space.
A semantic relevance map $r_t(x)\in[0,1]$ represents the available evidence associated with the target, while a Gaussian heatmap $b_t$ represents the robot's position. All five maps are spatially aligned $M\times M$ grids and are stacked into the neural spatial input:
\begin{equation}
O_t=[f_t,o_t,u_t,r_t,b_t]\in\mathbb{R}^{5\times M\times M}.
\label{eq:neural_observation}
\end{equation}
The current heading is supplied separately as $(\sin\theta_t,\cos\theta_t)$.

Temporal inputs are executed-pose history $H_t$ and the unexecuted previous plan $P_t$. Each contains up to 16 poses represented by
\begin{equation}
\boldsymbol{\xi}_i=
[x_i^{\mathrm{local}},y_i^{\mathrm{local}},
\sin(\Delta\theta_i),\cos(\Delta\theta_i),\rho_i].
\label{eq:history_value}
\end{equation}
Position and heading are relative to the current pose. Normalized sequence position $\rho_i$ spans $[-1,0]$ for history and $[0,1]$ for the previous plan. Both inputs are padded $16\times5$ arrays with validity masks, encoded separately from $O_t$.

\section{PROPOSED METHOD}

\subsection{System Overview}

SearchDiffusion (Fig.~\ref{fig:SystemOverview}) combines conditional map diffusion, deterministic dense utility prediction, and conditional trajectory diffusion. Observation-consistent map hypotheses and $O_t$ support utility prediction. Spatial features, predicted utility, current heading, and temporal context condition trajectory generation. Guidance refines candidates during denoising; validity checks assess their execution prefixes and a short lookahead before analytical selection.

With $M=128$, each planning step generates 32 candidates, each represented by 16 control tokens and decoded into $T=32$ heading-aware poses:
\begin{equation}
\tau_t^{(j)}
=\bigl[(x_i,y_i,\sin\theta_i,\cos\theta_i)\bigr]_{i=0}^{T-1}.
\label{eq:candidate_trajectory}
\end{equation}
The first pose matches the current robot pose. The robot executes
a prefix of the selected trajectory, updates observations and
temporal context, and replans.

\subsection{Counterfactual Map Hypotheses}
\label{subsec:counterfactual_maps}

The map generator samples $K$ complete layouts that preserve observed geometry while varying unknown regions. Its conditioning input is
\begin{equation}
C_t=[f_t,o_t,z_t]\in\mathbb R^{3\times M\times M},
\label{eq:map_condition}
\end{equation}
where $z_t=f_t+o_t$ marks observed cells.

A four-scale encoder (32/64/128/192 channels) conditions a U-Net with timestep-modulated residual blocks. The bottleneck uses a four-layer, four-head transformer (128-dimensional tokens on a $16\times16$ grid), 2D positional encoding, a global token, and a spectral residual block. Decoding predicts diffusion velocity and an auxiliary signed-distance field (SDF).

DDIM sampling starts from independent Gaussian noise and clamps predicted clean maps to known observations. Final samples are thresholded into binary maps $m_t^{(k)}$ (1: free, 0: occupied), then clamped again.

\begin{figure}[t]
\centering
\includegraphics[width=\columnwidth]{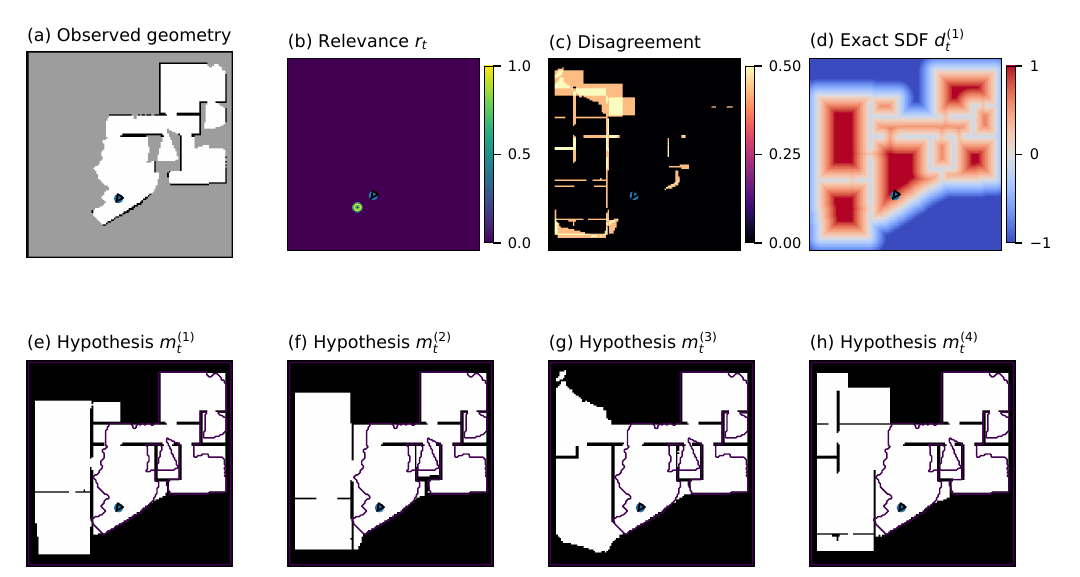}
\caption{Observed geometry (a), relevance input (b), and four map hypotheses (e--h). An exact SDF is computed for each hypothesis; (d) shows the first.}
\label{fig:hypothesis_and_sdf}
\end{figure}

For planning, SDFs $d_t^{(k)}$ are recomputed from binary geometry rather than taken from the auxiliary head. Positive values indicate free-space clearance and negative values indicate obstacle interiors; distances are clipped to $[-12,12]$ cells and divided by 12. The planner receives
\begin{equation}
\mathcal H_t=
\left\{(m_t^{(k)},d_t^{(k)})\right\}_{k=1}^{K},
\qquad K=4,
\label{eq:hypothesis_bank}
\end{equation}
with two aligned $M\times M$ grids per hypothesis. The generator is trained separately and frozen during planner training. Figure~\ref{fig:hypothesis_and_sdf} shows example outputs.

\subsection{Dense Utility Prediction}
\label{subsec:dense_utility}
For each training state, we compute a target map describing whether movement improves the search. Targets are defined on $\mathcal A_t$, the robot-connected subset of known-free cells in $f_t$. Let $a_t$ be the robot's rounded grid cell and $D_{\mathrm{geo}}(x,y)$ the shortest-path distance within this subset, using eight-neighbor moves without diagonal corner cutting.

The target combines frontier and semantic progress. Remaining search need $n_t\in[0,1]^{M\times M}$ marks cells that would benefit from further observation, and confidence $c_t\in[0,1]$ controls their relative importance. Training uses synthetic semantic observations (Sec.~\ref{subsec:training}); benchmark and physical-robot inference use $n_t=u_t$ and $c_t=\max_x r_t(x)$. These are search heuristics, not calibrated probabilities. 

For frontier progress, we sample up to two reachable viewpoints per frontier component and four headings per viewpoint. Let $\nu_t(x\mid v,\theta)$ be the fraction of hypotheses in which cell $x$ is visible from position $v$ and heading $\theta$, and let $\sigma_t(x)$ be the population standard deviation of binary free-space predictions, measuring disagreement. For a weighting map $w\in\{n_t,\sigma_t\}$, define
\begin{equation}
\begin{aligned}
B_w(v,\theta)&=1-\exp\!\left(-\frac{\sum_x\nu_t(x\mid v,\theta)w(x)}{180}\right),\\
Q_t(v)&=\max_\theta\frac{B_{n_t}(v,\theta)[1+B_{\sigma_t}(v,\theta)]}{2}.
\end{aligned}
\label{eq:frontier_viewpoint}
\end{equation}
Thus, disagreement boosts a viewpoint only when it reveals search need. Viewpoint quality is propagated through free space with a distance penalty and centered at the robot:
\begin{equation}
\begin{aligned}
W_t(x)&=\max_{v:Q_t(v)>0}\left[Q_t(v)-\frac{D_{\mathrm{geo}}(x,v)}{M}\right],\\
F_t(x)&=W_t(x)-W_t(a_t).
\end{aligned}
\label{eq:frontier_progress}
\end{equation}
Positive $F_t$ indicates improved access to useful viewpoints.

Semantic progress reduces the cost of reaching relevant evidence. The cost of reaching a destination cell $y$ combines path length plus a penalty $M(1-r_t(y))$, so more relevant destinations cost less:
\begin{equation}
\begin{aligned}
C_R(x)&=\min_{\substack{y\in\mathcal A_t\\r_t(y)>0}}
[D_{\mathrm{geo}}(x,y)+M(1-r_t(y))].
\end{aligned}
\label{eq:relevance_cost}
\end{equation}

\begin{figure}[t]
\centering
\includegraphics[width=\columnwidth]{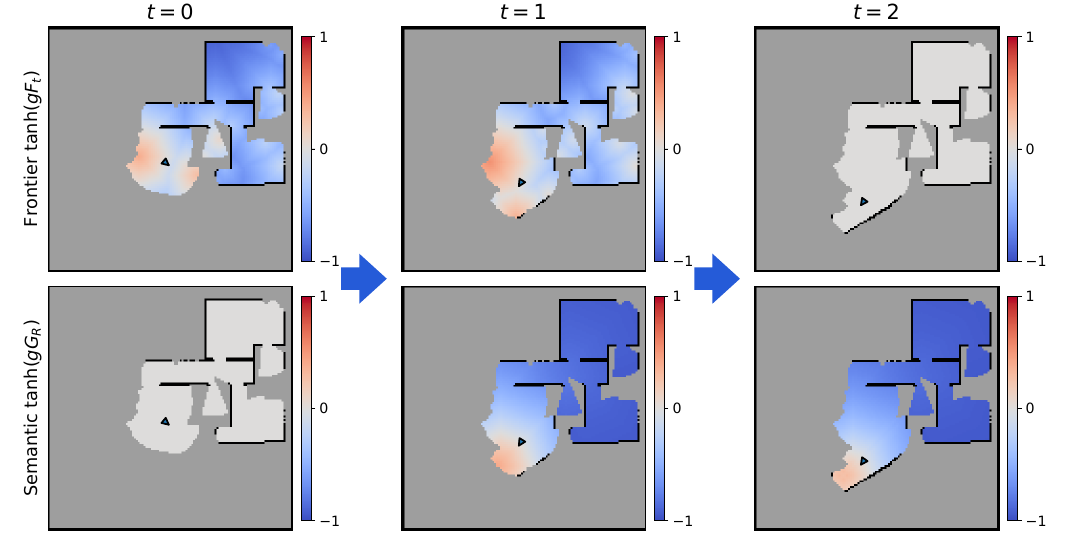}
\caption{Frontier (top) and semantic (bottom) progress fields at successive planning states. These are the components used to construct the utility target; they are not the final predicted utility map.}
\label{fig:utility_evolution}
\end{figure}

Moving to $x$ is progress when its remaining cost is lower than at the robot. Define this reduction as $\Delta C(x)=C_R(a_t)-C_R(x)$ and normalize it:
\begin{equation}
G_R(x)=\frac{\Delta C(x)}{M+|\Delta C(x)|}.
\label{eq:semantic_progress}
\end{equation}
The computed utility target is
\begin{equation}
V_t^{\mathrm{gt}}(x)=\frac{1-c_t}{2}\tanh(gF_t(x))
+\frac{1+c_t}{2}\tanh(gG_R(x)),
\label{eq:dense_utility_target}
\end{equation}
where $g=M/(2R_s)$ for sensor range $R_s$. The target lies in $[-1,1]$, is zero at $a_t$, and shifts from equal weighting at $c_t=0$ to semantic-only weighting at $c_t=1$ (Fig.~\ref{fig:utility_evolution}).
Convolutional encoders and order-invariant hypothesis attention produce spatial features $Z_t\in\mathbb R^{192\times16\times16}$. A three-stage upsampling decoder with a final $\tanh$ predicts $\hat V_t$. To condition trajectories, a three-layer strided convolutional encoder $E_V$ embeds the masked prediction:
\begin{equation}
\begin{aligned}
\widetilde V_t(x)&=\begin{cases}
\hat V_t(x),&x\in\mathcal A_t,\\
0,&x\notin\mathcal A_t,
\end{cases}\\
Z_t'&=Z_t+\gamma E_V(\operatorname{stopgrad}(\widetilde V_t)).
\end{aligned}
\label{eq:utility_conditioning}
\end{equation}
Here $\gamma$ warms up from zero to one. The stop-gradient operation preserves values but blocks trajectory-loss gradients through this feedback path. The backbone remains shared; predicted utility is used during training and inference.

\subsection{Conditional Trajectory Generation}
\label{subsec:trajectory_generation}
The model generates $L=16$ controls in the initial robot frame:
\begin{equation}
u_i=[\Delta x_i^{\mathrm{loc}}/s_p,\Delta y_i^{\mathrm{loc}}/s_p,
\sin\Delta\theta_i,\cos\Delta\theta_i],
\label{eq:control_representation}
\end{equation}
where $s_p=4$ is the position-normalization scale. Eight transformer blocks use 192-dimensional tokens and six attention heads. All blocks attend to $Z_t'$; alternating blocks also attend to history and the previous plan. Diffusion-step and heading context modulate the blocks through adaptive normalization.

A cubic B-spline decoder integrates the controls, preserves the current pose, incorporates measured initial motion, and resamples position and heading at shared locations into 32 poses. This encourages smoothness without imposing full vehicle dynamics. Inference uses ten DDIM steps, independent candidate noise, and cached conditioning features.

\subsection{Training Data and Objectives}
\label{subsec:training}
The map generator learns from HouseExpo~\cite{li2020houseexpo} with map splits preceding simulated observations and rotation/reflection augmentation. Denoising, reconstruction, and SDF objectives emphasize boundaries and narrow passages. After freezing it, a privileged teacher uses complete geometry to produce up to four diverse demonstrations per state. Planner pretraining starts at 3--10\% coverage; later training mixes these episodes equally with 30--90\% coverage episodes.

Synthetic relevance follows free-space distance to the target and is revealed only in observed cells. Detection raises confidence from zero to one, shifts relevance toward target approach, and sets search need to zero. Before detection, need is high in unknown cells and decreases with observations according to synthetic semantic uncertainty. Need and confidence construct utility targets and analytical scores, not separate neural inputs. Complete geometry and target coordinates remain restricted to simulation and the teacher.

The utility predictor and trajectory denoiser are trained jointly. For each training state, we sample one valid expert trajectory, favoring higher teacher scores. For a clean map (encoded in $[-1,1]$) or expert controls $X_0$, diffusion step $\kappa$ mixes signal and Gaussian noise:
\begin{equation}
\begin{aligned}
X_\kappa &= a_\kappa X_0+s_\kappa\epsilon,
\qquad \epsilon\sim\mathcal N(0,I),\\
v_\kappa &= a_\kappa\epsilon-s_\kappa X_0,\\
\hat X_0 &= a_\kappa X_\kappa-s_\kappa\hat v_\kappa.
\end{aligned}
\label{eq:training_noising}
\end{equation}
Here $\kappa$ indexes the diffusion steps, $a_\kappa=\sqrt{\bar\alpha_\kappa}$, and $s_\kappa=\sqrt{1-\bar\alpha_\kappa}$.

The planner objective combines denoising, utility prediction, and trajectory quality:
\begin{equation}
\mathcal L=\mathcal L_{\mathrm{denoise}}+
\mathcal L_{\mathrm{utility}}+\mathcal L_{\mathrm{trajectory}}.
\label{eq:total_planner_loss}
\end{equation}
Here $\mathcal L_{\mathrm{denoise}}$ is the mean Huber error against $v_\kappa$. The utility loss is the Huber error against $V_t^{\mathrm{gt}}$, averaged within $\mathcal A_t$ for each state. The trajectory loss is
\begin{equation}
\begin{aligned}
\mathcal L_{\mathrm{trajectory}}=\mathbb E\bigl[a_\kappa(&\ell_{\mathrm{quality}}\\
&+0.10\ell_{\mathrm{observed}}+0.08\ell_{\mathrm{hypothesis}}\\
&+0.05\ell_{\mathrm{clearance}})\bigr].
\end{aligned}
\label{eq:trajectory_loss}
\end{equation}
The expectation covers states, expert choice, noise, and diffusion steps; $a_\kappa$ weakens auxiliary penalties at high noise. The quality term combines expert-trajectory agreement, initial-motion continuity, previous-plan consistency, smoothness, and heading-change penalties.

The safety losses $\ell_{\mathrm{observed}}$, $\ell_{\mathrm{hypothesis}}$, $\ell_{\mathrm{clearance}}$ evaluate closely spaced points along the decoded
trajectory. Let $d_i^{(k)}$ be the normalized signed distance at
point $i$ in hypothesis $k$. 

At each point, hypothesis risk increases as clearance falls below
this margin. We average this risk across the $K$ hypotheses:
\begin{equation}
p_i^H=\frac{1}{K}\sum_{k=1}^{K}
\operatorname{clip}\left(1-\frac{d_i^{(k)}}{\mu},0,1\right).
\end{equation}
where we define a normalized clearance margin \(\mu=0.08\), corresponding to \(0.96\) grid cells. A hypothesis contributes zero risk when clearance is at least
$\mu$, and unit risk inside an obstacle. Observed-obstacle risk
$p_i^O$ is obtained by bilinearly sampling the known-occupied
mask $o_t$.

Pointwise risks are accumulated along the trajectory:
\begin{equation}
\begin{aligned}
\ell_{\mathrm{observed}}&=1-\prod_i(1-p_i^O),\\
\ell_{\mathrm{hypothesis}}&=1-\prod_i(1-p_i^H).
\end{aligned}
\end{equation}
These losses increase when more trajectory points are risky or
individual risks become larger. They are collision-risk
surrogates, not calibrated collision probabilities.

The additional clearance loss uses the largest predicted
clearance across hypotheses at each point:
\begin{equation}
\ell_{\mathrm{clearance}}
=\operatorname{mean}_i\left[
1-\exp\left(
-\left[1-\frac{\max_k d_i^{(k)}}{\mu}\right]_+^2
\right)\right].
\label{eq:training_clearance}
\end{equation}
where $[a]_+=\max(a,0)$. This term is zero whenever at least
one hypothesis provides sufficient clearance. It adds a
penalty where all hypotheses predict clearance below the margin.

Learning rate warms up over 5\% of the configured budget, remains constant until 80\%, then decays by cosine to 20\% of its initial value.

\subsection{Guidance, Safety, and Trajectory Selection}
\label{subsec:guidance_safety_selection}
During the final five denoising steps, guidance updates clean controls with weights and context fixed. Utility changes $\Delta V_E$ and $\Delta V_T$ compare the start with the execution-prefix and full-path endpoints, each truncated before first leaving $\mathcal A_t$. With observed-obstacle, hypothesis-clearance, and heading costs $C_O,C_H,C_\theta$, guidance maximizes
\begin{equation}
J_G=\underbrace{\Delta V_E+\tau_g\Delta V_T}_{\text{utility progress}}
-\underbrace{(4C_O+C_H+C_\theta)}_{\text{geometric and heading penalties}}.
\label{eq:guidance_score}
\end{equation}
The tail weight $\tau_g=0.25$ reduces emphasis on distant poses. For one-cell margin $m=1$, define $\phi(d)=[1-d/m]_+^2$. At dense position sample $p_i$, the geometric penalties are
\begin{equation}
\begin{aligned}
c_i^O
&=\phi\!\left(d_t^{\mathrm{obs}}(p_i)\right)
+\frac{\operatorname{dist}(p_i,\mathcal B)^2}{m^2},\\
c_i^H
&=\frac{1}{K}\sum_{k=1}^{K}
\phi\!\left(12d_t^{(k)}(p_i)\right).
\end{aligned}
\label{eq:local_guidance_costs}
\end{equation}
Here $d_t^{\mathrm{obs}}$ is signed distance from observed obstacles in cells, $\mathcal B=[0,M-1]^2$ is the map domain, and multiplication by 12 restores SDF cell units. Let $I_C$ include the execution prefix and four lookahead poses, and $I_R$ the remaining samples. For either cost sequence,
\begin{equation}
\mathcal R(c)=\tfrac12\bigl(\max_{i\in I_C}c_i+\operatorname{mean}_{i\in I_C}c_i\bigr)
+\tau_g\operatorname{mean}_{i\in I_R}c_i.
\label{eq:cost_reduction}
\end{equation}
We use $C_O=\mathcal R(c^O)$ and $C_H=\mathcal R(c^H)$, with an empty-tail mean of zero.  
Heading cost $C_\theta$ penalizes the squared excess of the wrapped heading--motion angle beyond $30^\circ$, divided by $\pi^2$, at both endpoints of moving segments. It equally combines the checked-prefix mean and root-mean-square (RMS) penalty, plus the weighted tail mean. Segments shorter than 0.05 cells are exempt, and this cost updates heading controls only.

Let $\eta$ contain position controls and angular corrections, and $g_\eta$ the gradient of $J_G$, with heading-alignment gradients restricted to angular controls. Guidance proposes
\begin{equation}
\eta^+=\Pi\!\left(\eta+\beta S\frac{Sg_\eta}{\max(\|Sg_\eta\|_\infty,10^{-9})}\right).
\label{eq:guidance_update}
\end{equation}
Diagonal $S$ sets 0.1-unit position and $10^\circ$ angle steps. Projection $\Pi$ limits changes to 0.25 units and $30^\circ$ while preserving the initial tangent. Backtracking tests $\beta=1,1/2,1/4,1/8$, accepting the first finite improvement above $10^{-7}$, otherwise leaving controls unchanged.

After decoding, validity checks reject nonfinite poses, inconsistent starts, known-obstacle intersections, and out-of-map samples over the execution prefix plus four lookahead poses. They also reject any checked point with clearance at most 0.02 (0.24 cells) in more than half the hypotheses.

The analytical selector counts each cell visible along the contiguous reachable path once. For this set $\Omega$, let $I_w=1-\exp(-\sum_{x\in\Omega}w_t(x)/180)$ for $w\in\{n,\sigma\}$, and $I_r=\sum_{x\in\Omega}r_t(x)/\max(\sum_x r_t(x),10^{-6})$. The score is
\begin{equation}
\begin{aligned}
J(\tau)={}&2I_n+1.5I_r+I_\sigma+2c_t r_t(p_{\mathrm{end}})\\
&+0.5D-10R-C_{\mathrm{motion}}+B_{\mathrm{explore}}.
\end{aligned}
\label{eq:analytical_selection_score}
\end{equation}
Here $p_{\mathrm{end}}$ is the last reachable position, $D$ the minimum along the full path of mean normalized hypothesis clearance, and $R$ the sum of known-obstacle and mean hypothesis-violation fractions. Motion cost $C_{\mathrm{motion}}$ weights path length, mean absolute curvature, RMS third position difference, mean heading change, and previous-plan directional disagreement by 0.025, 0.20, 0.04, 0.08, and 0.20, respectively.

The exploration bonus is
\begin{equation}
B_{\mathrm{explore}}=\frac{0.4F_s+0.3W_s+0.1E+0.08N-0.1Q}
{1+\exp((c_t-0.5)/0.1)}.
\label{eq:selection_exploration_bonus}
\end{equation}
$F_s$ measures usefulness-weighted geodesic progress toward frontiers. $W_s$ rewards new cells visible after turning one field-of-view width to either side at the endpoint, excluding cells already visible along the path. It uses the saturating information score with weights $(n_t+r_t+\sigma_t+u_t)/4$. Efficiency $E$ is endpoint displacement divided by reachable path length (zero for a stationary path). Novelty $N$ is the endpoint's distance to the nearest recorded history position, divided by 16 cells and clipped to $[0,1]$ (zero for empty history). The low-efficiency penalty is $Q=[(0.35-E)/0.35]_+$. Analytical features determine ranking; predicted utility influences generation and guidance. The highest-scoring valid candidate is selected; if none passes the checks, no trajectory is issued.

\subsection{Robot-Specific Execution Constraints}

A holonomic controller tracks position and body heading, including lateral and reverse motion. Fresh observations revalidate the remaining prefix; local swept-footprint and stopping checks can halt execution. Map invalidation may trigger bounded replanning. Footprint constraints are enforced during execution, not learned.

\begin{figure}[t]
\centering
\includegraphics[width=0.7\columnwidth]{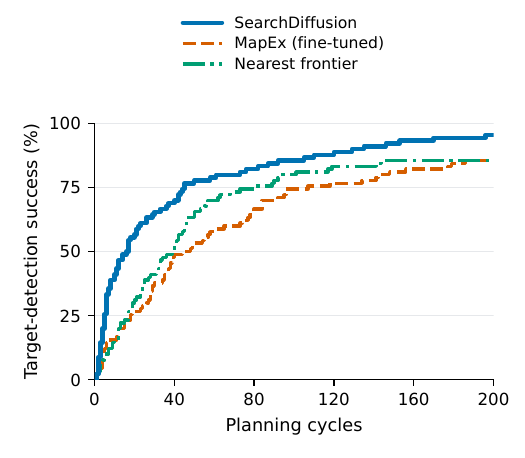}
\caption{Target-detection success versus planning cycles for the compared planners.}
\label{fig:baseline_comparison}
\end{figure}

\begin{figure}[t]
\centering
\includegraphics[width=\columnwidth]{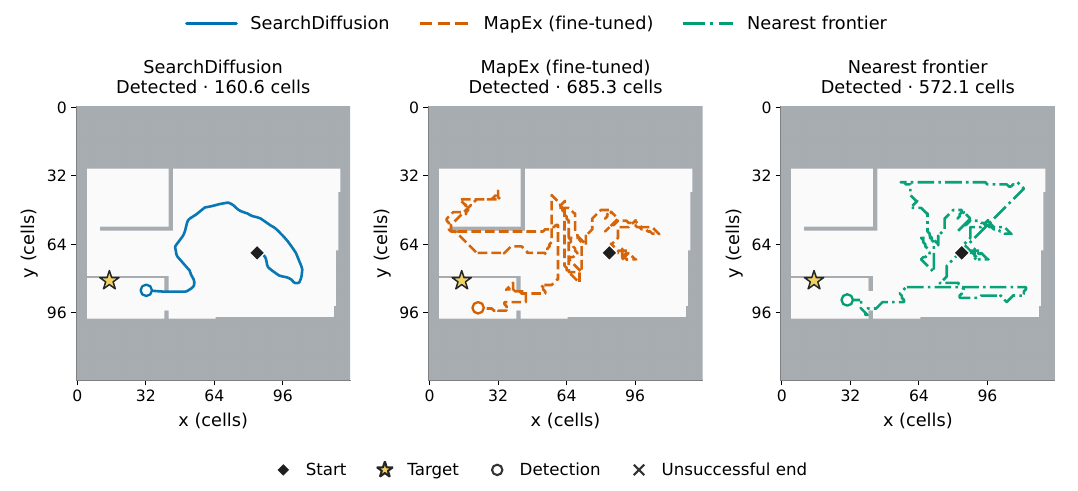}
\caption{Executed search paths on one matched task for SearchDiffusion, fine-tuned MapEx, and nearest frontier. Each path combines motion across successive replanning cycles; these are not individual diffusion candidates.}
\label{fig:example_trajs}
\end{figure}

\begin{figure}[t]
\centering
\includegraphics[width=0.7\columnwidth]{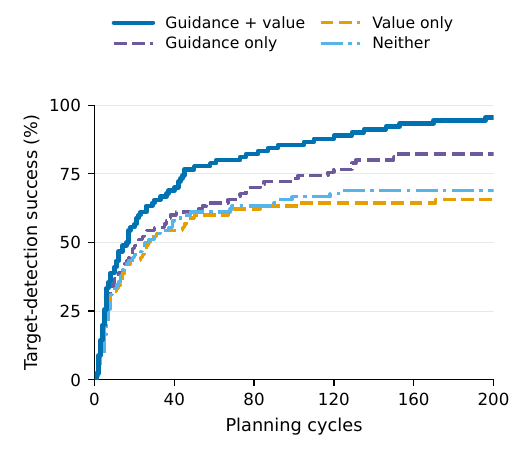}
\caption{Target-detection success versus planning cycles for the four utility–guidance ablations.}
\label{fig:ablation_compare}
\end{figure}

\begin{figure}[t]
\centering
\includegraphics[width=\columnwidth]{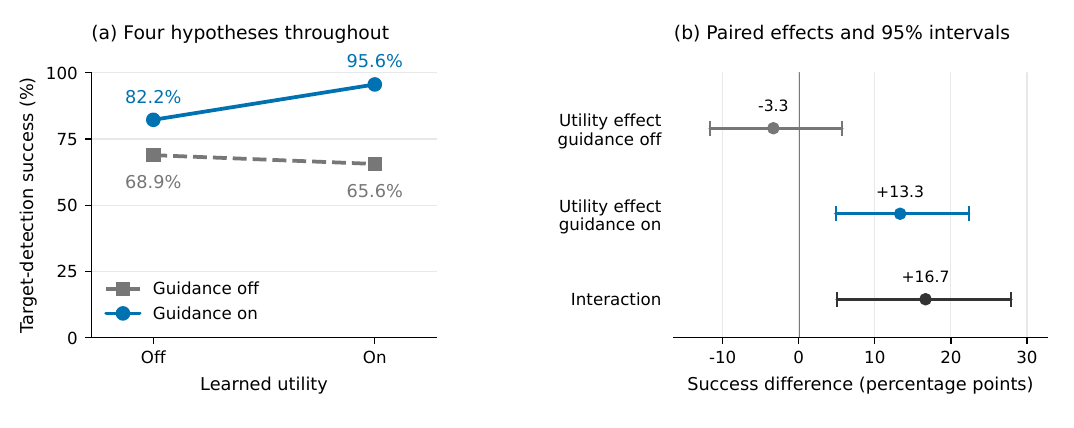}
\caption{The utility–guidance interaction.}
\label{fig:utility_guidance}
\end{figure}

\begin{table*}[t!]
\caption{Baseline comparison on 90 tasks: estimates [95\% confidence intervals].}
\label{tab:baseline_comparison_ci}
\centering
\begingroup
\setlength{\tabcolsep}{2pt}
\renewcommand{\arraystretch}{1.0}
\begin{tabular*}{\textwidth}{@{\extracolsep{\fill}}lcccc@{}}
\toprule
\textbf{Method} & \textbf{Detection (\%)} $\uparrow$ & \textbf{AUC} $\uparrow$ & \textbf{Distance (cells)}$^{\dagger}$ & \textbf{Time (ms)} $\downarrow$ \\
\midrule
SearchDiffusion & \textbf{95.6}~[88.4, 100.0] & \textbf{0.803}~[0.734, 0.872] & 78.2~[45.0, 116.7] & 307.3~[299.0, 325.7] \\
MapEx (stock) & 83.3~[73.3, 92.3] & 0.607~[0.512, 0.695] & 206.1~[161.7, 269.2] & 45.2~[44.6, 45.7] \\
MapEx (fine-tuned) & 85.6~[75.3, 94.3] & 0.639~[0.542, 0.729] & 177.9~[135.3, 240.0] & 44.6~[44.2, 45.4] \\
Nearest frontier & 85.6~[75.0, 94.8] & 0.696~[0.598, 0.780] & 150.5~[101.7, 191.2] & \textbf{0.86}~[0.76, 0.97] \\
\bottomrule
\end{tabular*}
\endgroup
\par\smallskip
\begin{minipage}{\textwidth}

\end{minipage}
\end{table*}

\begin{table*}[t!]
\caption{Frozen-checkpoint ablations on the same tasks: estimates [95\% confidence intervals].}
\label{tab:ablation_comparison_ci}
\centering
\begingroup
\setlength{\tabcolsep}{2pt}
\renewcommand{\arraystretch}{1.0}
\begin{tabular*}{\textwidth}{@{\extracolsep{\fill}}lccccccc@{}}
\toprule
$K$ & \textbf{U} & \textbf{G} & \textbf{Detection (\%)} $\uparrow$ & \textbf{AUC} $\uparrow$ & \textbf{Distance (cells)}$^{\dagger}$ & \textbf{Time (ms)} $\downarrow$ \\
\midrule
 4 & On & On & \textbf{95.6}~[88.4, 100.0] & 0.803~[0.734, 0.872] & 78.2~[45.0, 116.7] & 307.3~[299.0, 325.7] \\
 1 & On & On & 93.3~[87.0, 98.9] & 0.819~[0.744, 0.891] & 71.6~[34.3, 113.3] & 293.9~[282.8, 311.1] \\
 0 & On & On & 94.4~[90.2, 98.0] & \textbf{0.823}~[0.763, 0.879] & 77.0~[33.5, 119.0] & 235.7~[221.4, 265.4] \\
 0 & Off & On & 85.6~[77.6, 92.9] & 0.728~[0.629, 0.818] & 51.8~[37.8, 82.0] & 127.0~[125.9, 128.6] \\
 4 & On & Off & 65.6~[57.1, 73.9] & 0.588~[0.501, 0.674] & 54.2~[32.6, 81.8] & 151.6~[150.9, 152.1] \\
 4 & Off & Off & 68.9~[60.0, 78.2] & 0.612~[0.524, 0.700] & 57.3~[37.5, 87.2] & 146.2~[145.4, 146.7] \\
 0 & Off & Off & 62.2~[48.9, 73.8] & 0.578~[0.456, 0.690] & 40.0~[30.0, 72.9] & \textbf{70.0}~[69.5, 70.6] \\
 4 & Off & On & 82.2~[73.8, 90.6] & 0.687~[0.605, 0.777] & 71.5~[40.4, 121.0] & 212.1~[210.7, 214.9] \\
\bottomrule
\end{tabular*}
\endgroup
\par\smallskip
\begin{minipage}{\textwidth}

\end{minipage}
\end{table*}

\section{EXPERIMENTS AND RESULTS}
\subsection{Experimental Setup and Evaluation Protocol}

We evaluated 90 tasks across 24 maps using one inference seed, with matched initial poses, observed areas, targets, and sensing conditions. The simulated sensor had an 18-cell range, a $60^\circ$ field of view, and obstacle occlusion. Runs allowed 200 planning cycles, executing up to ten poses per cycle. SearchDiffusion generated 32 candidates with ten denoising steps per decision.

Targets started undiscovered; runs ended at first detection. Success curves retain failures in the denominator and use planning cycles or travelled distance as budgets. Success--cycle AUC averages success over cycle budgets 1--200. Planning latency included map prediction, generation, guidance, safety checks, and selection, measured on 32 shared initial states on one workstation (RTX~4080 for learned models).

Excluding 51 tasks overlapping MapEx's fine-tuning validation maps leaves 39 tasks across nine maps with no detected fine-tuning overlap.

We report 95\% percentile confidence intervals from 2,000 map-cluster bootstrap resamples, drawing maps with replacement and retaining all their tasks. Shared resamples preserve paired method comparisons and ablation contrasts. Intervals do not capture inference-seed variability and are unadjusted for multiple comparisons.

Tables~\ref{tab:baseline_comparison_ci} and~\ref{tab:ablation_comparison_ci} report median planning times and median distances over each method's successful episodes. Successful subsets differ, so shorter distances alone do not establish greater efficiency.

\subsection{Comparison with Baseline Planners}

Baselines included MapEx with released predictors or predictors fine-tuned for five epochs on HouseExpo, and nearest-frontier planning, which selects nearby boundaries of unexplored space. SearchDiffusion achieved the highest detection rate and success--cycle AUC, at greater planning cost (Table~\ref{tab:baseline_comparison_ci}).

SearchDiffusion, fine-tuned MapEx, and nearest frontier recorded no executed-path obstacle contacts; their median per-episode minimum clearances were 0.71, 0.50, and 0.50 cells. On the restricted subset, they detected 38/39, 36/39, and 37/39 targets, respectively, indicating smaller performance differences.

\subsection{Component Ablation Studies}

Eight variants changed the number of map hypotheses, utility, and guidance while keeping network weights fixed. Guidance included geometric and heading refinement; validity checks remained active throughout.

Utility helped most with guidance (Fig.~\ref{fig:utility_guidance}). With four hypotheses, adding utility increased success from 82.2\% to 95.6\% when guidance was enabled. Without guidance, success changed from 68.9\% to 65.6\%; this $-3.3$-point difference had an interval of $[-11.7,5.7]$, providing no clear evidence of harm. The difference between utility effects was $+16.7$ percentage points, with an interval of $[5.0,27.9]$.

With utility enabled, guidance reduced failures due to no valid candidate from 26 to two. With guidance enabled, utility reduced budget-exhaustion failures from 15 to two, suggesting complementary roles in executable motion and productive search.

With utility and guidance enabled, four, one, and zero hypotheses produced 86, 84, and 85 detections (Table~\ref{tab:ablation_comparison_ci}). The four-versus-zero difference was $+1.1$ points, with an interval of $[-5.1,6.5]$. Removing hypotheses reduced median planning latency from 307 to 236~ms (approximately 23\%), suggesting a lower-cost configuration.

\subsection{Real-World Evaluation}

We deployed SearchDiffusion on a ZSL-1 quadruped with a Livox MID-360 LiDAR/IMU and an Intel RealSense D455 RGB-D camera. GLIM~\cite{koide2024glim} mapping and trajectory execution run onboard a 16-GB Jetson Orin NX, while the diffusion models and OneMap run on an external RTX 4090 workstation. Both diffusion models also support onboard inference: warmed-up recorded-input tests without ROS or GLIM required approximately 3.6--3.8 s per plan, including map-hypothesis generation.

The planner receives a $128\times128$ occupancy grid at 0.1 m resolution, accumulating observations from a $60^\circ$ forward sector with a 4 m range. This restriction applies only to planner observations. OneMap asynchronously processes RGB-D inputs at up to 2 Hz to generate relevance maps for a user-specified text query. The robot executes up to 16 subsequent poses per 32-pose plan, with speed limits of 0.20 m/s and 0.50 rad/s. Independent local-costmap and swept-footprint checks supervise execution. Pretrained weights were deployed without real-robot fine-tuning. Robot-specific settings used a soft guidance clearance margin of 0.65 m and a hard minimum-clearance threshold of 0.20 m for trajectory validation. Across ten physical object-search environments, success was approximately 80\%, with unsuccessful trials attributed to false positives in the relevance maps.

\begin{table}[t]
\centering
\caption{Computation times by platform and recorded robot-run performance.}
\label{tab:deployment_latency}
\small
\begin{tabular*}{\columnwidth}{@{\extracolsep{\fill}}lccc@{}}
\toprule
\multicolumn{4}{@{}l}{\textbf{Computation time by platform (s)}} \\
\midrule
Device & Map gen. & Planning & Total \\
\midrule
Jetson Orin NX & 1.08--1.09 & 2.49--2.74 & 3.57--3.82 \\
GeForce MX450  & 1.57--1.58 & 1.11--1.21 & 2.67--2.79 \\
RTX 4090      & 0.32--0.35 & 0.30--0.34 & 0.63--0.69 \\
\midrule
\multicolumn{4}{@{}l}{\textbf{Recorded robot runs}} \\
\midrule
\multicolumn{3}{@{}l}{Success rate}
    & 8/10 \\
\multicolumn{3}{@{}l}{Minimum obstacle distance (m)}
    & 0.26 \\
\multicolumn{3}{@{}l}{Median replans}
    & 7.50 \\
\multicolumn{3}{@{}l}{Median episode time (s)}
    & 79.61 \\
\bottomrule
\end{tabular*}
\end{table}

Table~\ref{tab:deployment_latency} reports computation-time ranges across deployment platforms. Planning includes guidance, validity checks, and selection; network communication and OneMap are excluded.

\begin{figure}[t!]
\centering
\includegraphics[width=\columnwidth]{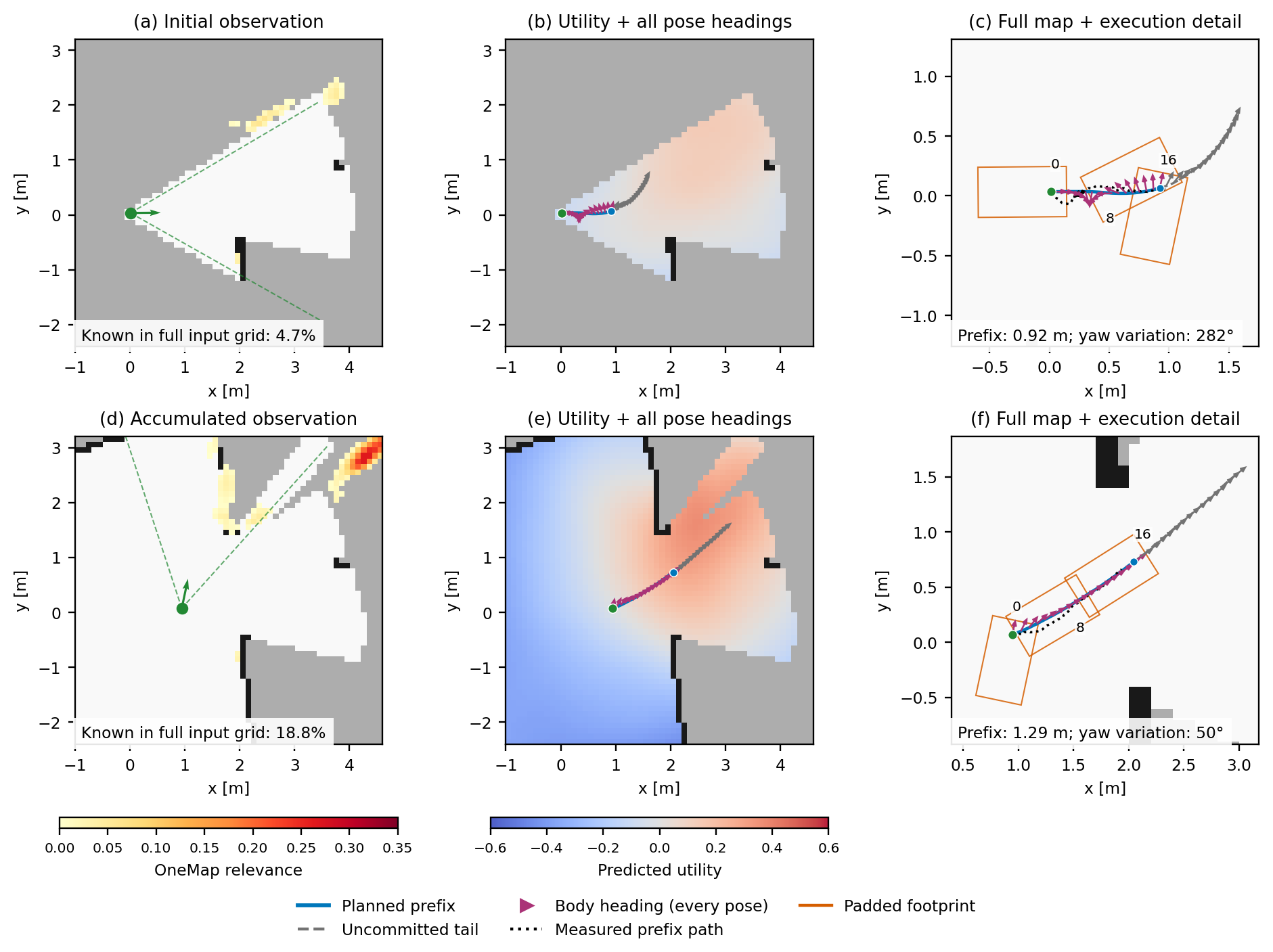}
\caption{Position and body-heading plans under successive observations.}
\label{fig:heading_observation}
\end{figure}

Figure~\ref{fig:heading_observation} shows consecutive plans from one physical run. With sparse initial observations, the first 0.92~m prefix changes body heading by approximately $282^\circ$. By the next decision, observed grid coverage increases from 4.7\% to 18.8\%. The next prefix spans 1.29~m, with mean heading--translation deviation decreasing from $86^\circ$ to $11^\circ$. These examples illustrate independent planning of translation and viewing direction, without establishing that rotation is optimal or caused solely by map coverage.

\section{CONCLUSIONS}

SearchDiffusion combines learned dense utility with diffusion planning for object search under partial observability. The utility map conditions trajectory generation and guides refinement, followed by validity checks and analytical selection. In matched simulations, the method achieved higher detection success across planning budgets than the tested frontier baselines, at greater computational cost. Deployment on a quadruped achieved an operator-assessed success rate of 80\% without fine-tuning on real-robot data.

Ablations indicate complementary benefits: guidance reduced invalid-trajectory failures, while utility reduced failures caused by exhausting the search budget. Runtime map hypotheses provided limited additional detection benefit; removing them reduced computation while maintaining similar observed success. These findings support combining learned search utility with guided trajectory generation.

Future work will improve onboard planning efficiency, incorporate the robot's physical footprint and sensing uncertainty into trajectory generation, and evaluate performance in a broader range of simulated and real environments.

\bibliographystyle{IEEEtran}
\bibliography{references}

\end{document}